# Blockchain Transaction Fee Forecasting: A Comparison of Machine Learning Methods

Conall Butler [1] and Martin Crane [1,2,*]

[1] School of Computing, Dublin City University, Glasnevin, Dublin 9, Ireland; conall.butler36@mail.dcu.ie
[2] ADAPT Research Centre, Dublin City University, Glasnevin, Dublin 9, Ireland
* Correspondence: martin.crane@dcu.ie

**Abstract:** Gas is the transaction-fee metering system of the Ethereum network. Users of the network are required to select a gas price for submission with their transaction, creating a risk of overpaying or delayed/unprocessed transactions involved in this selection. In this work, we investigate data in the aftermath of the London Hard Fork and shed insight into the transaction dynamics of the network after this major fork. As such, this paper provides an update on work previous to 2019 on the link between EthUSD/BitUSD and gas price. For forecasting, we compare a novel combination of machine learning methods such as Direct-Recursive Hybrid LSTM, CNN-LSTM, and Attention-LSTM. These are combined with wavelet threshold denoising and matrix profile data processing toward the forecasting of block minimum gas price, on a 5-min timescale, over multiple lookaheads. As the first application of the matrix profile being applied to gas price data and forecasting that we are aware of, this study demonstrates that matrix profile data can enhance attention-based models; however, given the hardware constraints, hybrid models outperformed attention and CNN-LSTM models. The wavelet coherence of inputs demonstrates correlation in multiple variables on a 1-day timescale, which is a deviation of base free from gas price. A Direct-Recursive Hybrid LSTM strategy is found to outperform other models, with an average RMSE of 26.08 and $R^2$ of 0.54 over a 50-min lookahead window compared to an RMSE of 26.78 and $R^2$ of 0.452 in the best-performing attention model. Hybrid models are shown to have favorable performance up to a 20-min lookahead with performance being comparable to attention models when forecasting 25–50-min ahead. Forecasts over a range of lookaheads allow users to make an informed decision on gas price selection and the optimal window to submit their transaction in without fear of their transaction being rejected. This, in turn, gives more detailed insight into gas price dynamics than existing recommenders, oracles and forecasting approaches, which provide simple heuristics or limited lookahead horizons.

## 1. Introduction

Blockchain technologies and their applications such as cryptocurrencies, smart contracts, Non-Fungible Tokens (NFTs) and DeFi (Decentralized Finance) show great potential for disruption and innovation, and they are much discussed. The development of these decentralized applications is enabled through the Ether cryptocurrency, the associated blockchain Ethereum, and the Ethereum Virtual Machine. Ether (ETH) is the second largest cryptocurrency by market cap after Bitcoin. Use of the Ethereum network is growing; daily transactions rose from 500,000 to 2,000,000 between 2018 and 2023 [1].

Ethereum network transactions are cryptographically signed instructions between accounts. These instructions can be as simple as a transfer of ETH or more complex contract deployments that enable a variety of decentralized applications. Gas is the unit of computational work used when processing a transaction on the network. The number of gas units consumed by a transaction is dependent on the computational complexity of the transaction. Gas has a price per unit in ETH, and the price is submitted by the sender with the transaction [2]. The process of packing transactions into blocks proceeds as follows: many transactions can go into a single block in Ethereum with miners carrying out a number of tasks:

The list of pending transactions arranged by gas price—and hence processing priority—is the first parameter that the miners have to work with. In addition, the amount of transactions that miners can add to a block is restricted. After the miners have decided which transactions should be packed, the Proof of Work procedure starts [3]. If the total amount of gas used by all the transactions is greater than the block's upper limit, the block will not be recognized by the Ethereum network. If this is not the case, the transaction can be included in the block and the associated reward is given to the miner who finds the new block first. The selection of transactions by miners has been shown to be almost exclusively based on the submitted gas price [4].

There is risk associated with gas price selection when submitting a transaction; too high will result in unnecessarily high fees, while selecting too low can incur transaction wait times or failure of the transaction to be processed if not selected by miners. High gas fees are seen as a major impediment to applications on the Ethereum network. The impact of gas fees on applications can be seen in cases such as ConstitutionDAO [5].

It was in part to address such issues that the Ethereum London Hard Fork was introduced on 5 August 2021 [6]. One innovation introduced here is the move from Proof of Work to Proof of Stake. The main motivation behind this was rather than processing power being used for voting, users become validators on the basis of the number of staked coins they have. Proof of Stake is designed to allow for better energy efficiency and a lower bar for entry [7].

Prior to the introduction of Ethereum 2.0 (Serenity) [8] and the switch to the Proof of Stake system, it was necessary to make several preparations, and these were introduced in the five further Ethereum Improvement Proposals (EIPs) [9]. One of these, EIP-1559, was designed to make fees more user-friendly and increase the uniformity of the transactions mined in a block. Additionally, this proposal is aimed at reducing overpayments to miners. EIP-1559 offers a variable block size with a 50% target usage. Thus, the majority of the time slots will be only halfway full. There may still be spikes when there are full blocks for a while, but it is more likely to happen for brief intervals. As the dataset used for this paper post-dates this introduction, we cover the details of EIP-1559 briefly below.

Several gas price recommenders (or oracles) exist to aid the gas price prediction task currently. These recommenders use simple heuristics and past data to generate a number of recommendations. Go-Ethereum (Geth) recommends a gas price to submit for the next block based on a percentile of minimum block gas prices for the past number of blocks, defaulting to the 60th percentile of the last 20 blocks [10]. EthGasStation estimates the number of blocks waited when a transaction is submitted at a specified gas price, which is based on a Poisson regression model using the previous 10,000 blocks of data [11]. GasStation—Express estimates the likelihood of a transaction being included in the next block at a gas price based on proportion of the last 200 blocks with a transaction at that price or lower [12]. The performance of these oracles has, however, not lived up to expectations in many cases (as will be detailed below) [13].

This paper is related to previous gas price forecasting and recommender work by Mars et al. [14] and Werner et al. [15]. The aim of this study is to first investigate the relation between potential model inputs in blockchain and exchange data, using wavelet coherence as seen Garrigan et al. [16], Sun and Xu [17] and Qu et al. [18]. The next stage is development of a forecasting model based on these inputs. Previous approaches have applied Long Short-Term Memory (LSTM) and Attention-Gated Recurrent Unit (GRU) models [3]. This study intends to investigate performance over different forecast horizons, using multiple approaches; a direct-recursive hybrid LSTM forecasting approach, inclusion of an attention mechanism with the matrix profile (as seen applied to low-granularity daily COVID data by Liu et al. [19]) and also Convolutional Neural Networks (CNNs) fed to LSTM architectures, or CNN-LSTMs. A comparison of these methods has been made recently by Chandra et al. [20].

Wavelet denoising will also be investigated, as seen in Dyllon et al. [21] and Qiu et al. [22]. A combination of wavelet transforms, matrix profile and attention-LSTM methods toward time-series forecasting is a novel approach to our knowledge, particularly in the domain of blockchain transaction fees.

We feel that our paper contributes to the literature through:

1. First and foremost, the time period studied is in the aftermath of the so-called Ethereum London Hard Fork when the immediate aftereffects of this had passed. In particular, we feel that Research Question 3 of our study provides an update on Pierro and Rocha's work of 2019 [23] on the link between EthUSD/BitUSD and gas price.
2. This study is the first that we have found to investigate performance over different forecast horizons. These time horizons are useful, as a user must select between these and potentially be penalized in terms of cost or missed transactions for choosing one over the other. There is thus a real cost penalty for the user in not choosing correctly here.
3. In our study, we use multiple approaches: a direct-recursive hybrid LSTM forecasting approach, inclusion of an attention mechanism with the matrix profile, as seen applied to low-granularity daily COVID data and also Convolutional Neural Networks (CNNs), fed to LSTM architectures, or CNN-LSTMs. In the case of matrix profiles, this is the first incidence that we could find of the use of the method in gas price prediction.

These, we feel, provide an academic and practical justification for why this research is warranted at the current time. Specifically, the Research Questions and aims of this paper are as follows, to be addressed using data from 26 November 2021 to 31 April 2022:

**RQ1.** What is the best method to forecast minimum block price across multiple lookaheads, comparing several modeling approaches?

**RQ2.** Wavelet transforms and the matrix profile are unstudied methods in this area; can these methods improve forecasting metrics or provide insight into gas price mechanics?

**RQ3.** How do blockchain and ETH cryptocurrency exchange data relate to gas price, and can these data be used to improve forecasting metrics?

The sections contained in this paper are: Section 2. Glossary; Section 3. Gas Mechanics Literature Survey; Section 4. Previous Work on Gas Price Prediction; Section 5. Materials and Methods; Section 6. Methods for Data Modeling; Section 7. Results; Section 8. Discussion; Section 9. Conclusions.

**2. Glossary**

*Ethereum Network Terminology [4]*

- Block: Batch of transactions added to the blockchain.
- Contract/Smart Contract: Complex transaction, with clauses and dependencies for operation; not a simple transfer of ETH. Basis of complex applications.
- ETH: Ether, cryptocurrency of the Ethereum network.
- Gas: Unit of computational work completed when processing transaction on the Ethereum network. The gas required to process transactions increases with transaction complexity.
- Gas Price: Fee paid to miners by transaction sender, per unit of gas, to process a transaction and include it in the blockchain. Operates on priority queuing basis: the highest gas price transactions are selected by miners, the gas price is selected by transaction senders. Price is typically quoted in gwei.

- Gwei: The denomination of ETH cryptocurrency. One ETH is equivalent to 1018 wei. A giga-wei, or gwei, is equivalent to $10^9$ wei, or $10^{-9}$ ETH. All gas price values given in this work are in gwei.
- Mempool: Cryptocurrency nodes that function as a way to store data on unconfirmed transactions, acting as a transaction waiting room prior to inclusion in a block.
- Miner: Third party that performs necessary computations for the inclusion of transaction on the blockchain, at a fee.
- Transaction: Cryptographically signed instruction from one Ethereum network account to another, which includes simple ETH transfer and more complex contract deployments that allow for various applications on the network.

## 3. Gas Price Mechanics Literature Survey

### 3.1. Economics of Ethereum Gas Price

Economic determinants of gas price based on blockchain and cryptocurrency exchange data are investigated by Donmez et al. [24]. A strong non-linear association is found between block utilization and both marginal and median daily gas prices. Gas price is found to be highly influenced by block utilization above 90%, with minimal impact below 90%. ETH transfer transactions are found to be more urgent than smart contact transactions, and a higher proportion of transfers is found to be associated with higher gas prices. Gas price is found to be negatively associated with ETH value. This is consistent with the principle of network users being concerned with network usage costs in term of real currency value [23].

The inclusion of transactions in the next mined block operates on a priority queuing mechanism and is shown to comply with economic predictions from queueing theory and supply/demand theory [23]. Basic ETH transfer-type transactions are observed to have higher urgency and thus typically higher gas price submission. This is because miners select transactions for inclusion based almost solely on gas price [5]. It is assumed that the observed minimum gas price variable will begin to rise when sufficient numbers of high-priority, higher paying transactions are available to fill mining capacity, and transactions close to the base fee are no longer selected. We can observe that the min-gas price rarely deviates from the base fee; however, cases do occur where there is significant deviation. It is possible that lower and upper percentiles of gas prices within blocks may contain some predictive information as to these events, as mining capacity is gradually filled.

The block base fee, the minimum gas price for a submitted transaction in order for it to be eligible for inclusion in the block, is related to block size through a process known as *tâtonnement*. Blocks have a target size of 15 million gas, and the size is adjusted to meet network demands up to a maximum of 30 million gas worth of transactions per block. The base fee is increased by up to 12.5% of the previous blocks base fee, when the previous block is above the target, continually increasing until the block size has returned to the target [4]. The process from transaction submission to inclusion in the blockchain is shown in Figure 1.

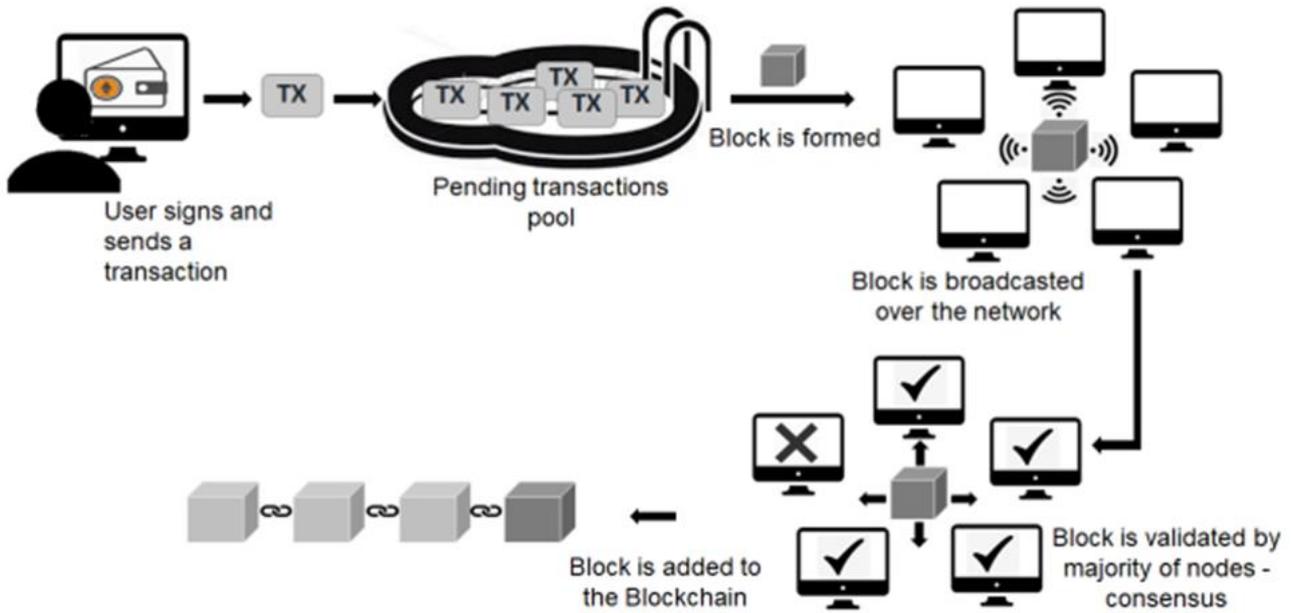

**Figure 1.** Ethereum Blockchain Flow (reproduced with permission from Mars et al. [14]).

*3.2. Influencing Factors on Ethereum Gas Price*

Influencing factors of gas price, and also the reliability of gas price oracles, are investigated by Pierro et al. [24]. Ref. [24] defines gas price as the Etherchain "Fast" price, which is defined as the price where 90% of the previous 200 mined blocks contain a transaction at this price. Transactions submitted at this price are expected to be processed by miners within 1–2 min. The gas price is indicated to have pairwise Granger causation with miner count and unconfirmed transaction count at $p$ = 0.05. Both cases have negative Pearson correlation. Gas price was not found to share Granger causality with the other tested variables: hash rate, bloc time, block difficulty, ETH/US Dollar, and ETH/Bitcoin. Strangely, although some (Werner et al. [15] and Liu et al. [25]) have used the ETH price as an input for their models, and Donmez et al. [24] also talk about a negative association between ETH and gas price, this current work is the first we have found to investigate the relationship between them as opposed to just modeling based on the ETH price. It is to address this—particularly in light of EIP-1559—that we relook at this issue here.

Liu et al. also looked at influencing factors on gas price [25]. They present a Machine Learning Regression (MLR)-based approach to predicting gas prices with the goal of locating the next block's lowest transaction gas price for conducting cost-effective Ethereum transactions. Specifically, they identify five influencing parameters from the Ethereum transaction process (i.e., difficulty, block gas limit, transaction gas limit, ether price, and miner reward) and use a traditional machine learning regression to develop the predictive model. The proposed MLR technique appears to function effectively and can lead to considerable potential savings for all transactions with a 74.9% accuracy, according to their empirical analysis on 194,331 blocks.

*3.3. Experiences around The Ethereum Hard Fork*

The lack of clarity around Ethereum gas fees was in part the reason that Ethereum-London Hard Fork was introduced on 5 August 2021. Prior to the introduction of Ethereum 2.0 (Serenity) and the switch to the Proof of Stake system (PoS), it was necessary to make several preparations, and these were introduced in the five further Ethereum Improvement Proposals (EIPs). The element of the Ethereum protocol that establishes the cost for every transaction added to the blockchain is the transaction fee mechanism. Historically, Ethereum employed a first-price auction fee mechanism. EIP-1559 suggested making several changes to this, e.g., introducing variable-size blocks, a history-dependent reserve price, and the burning of a large part of the transaction fees, to conserve the value of the currency [26]. EIP-1559's influence on user experience and market performance in the immediate aftermath of its launch was assessed by Reijsbergen et al. [27] using on-chain data. Empirical results indicate that although EIP-1559 generally succeeds in achieving its objectives, its short-term behavior is characterized by severe, chaotic oscillations in block sizes (as predicted by the authors' most recent theoretical dynamical system analysis) and sluggish adjustments during demand spikes (such as NFT drops). Unwanted inter-block fluctuation in mining rewards is caused by both occurrences. An alternate base fee adjustment method is suggested that uses an additive increase, multiplicative decrease (AIMD) updating strategy to account for this. Simulations demonstrate that under various demand scenarios, the latter robustly beats EIP-1559. Results show that variable learning rate methods may be a viable alternative to EIP-1559, advancing ongoing talks on the creation of transaction fee marketplaces with higher levels of efficiency.

Liu et al. also looked at the impact at the impact of the introduction of EIP-1559 [28] using the available data from the Ethereum blockchain, the Mempool, and exchanges. To investigate its causal impact on blockchain transaction cost patterns, transaction waiting times, and consensus security, they found that EIP-1559 enhances user experience by minimizing intra-block variations in gas prices paid and cutting down on user wait times. However, they also discover that waiting time is substantially longer when Ether's price is more erratic.

Lan et al. [29] propose a machine learning-based method to forecast the gas price of upcoming blocks paired with a dynamic feature also explored from Mempool. In particular, they took into account pending transactions and their gas cost in the Mempool and used them for the first time as a machine learning feature. For prediction, they mix the Mempool features with machine learning models with results showing good prediction ability, especially in the two indices MAE and RMSE.

## 4. Previous Work on Gas Price Prediction

### 4.1. The Role and Performance of Gas Price Oracles

The role of gas price recommenders or oracles in the prediction of the gas price has been discussed by a number of authors [30–34]. In brief, the gas price oracle attempts to predict the future gas price on the basis of previous block utilization. If the oracle indicates a lower than 100% utilization, this tends to show that there was spare capacity and hence there could be an opportunity to reduce gas price bid. Conversely, utilization at more than 100% would indicate that a reduced bid would incur the risk of its transaction not being selected by the miners. To help set the right gas price, the Gas Oracle categorizes the gas price into categories based on the interval of time the user might be willing to wait and for each of them suggests a gas price to set [31].

Empirical analysis of historic gas price data, proposal of a gas price recommendation algorithm and driving GRU-network based gas price forecast can be seen in Werner et al. Implementing an additional wait time of 4.8 blocks (~60 s) with the proposed approach resulted in a saving of 75% on gas fees when compared to the popular Go-Ethereum (Geth) recommender. Forecast evaluation metrics are not discussed. The recommendation algorithm was fed ground truth gas price data, which showed further improvement on the GRU-driven forecast, indicating room for improvement on the forecasting model. Empirical analysis of the gas price data shows high volatility with mean maximum gas price exceeding mean minimum gas price by orders of magnitude and the average block gas price having a mean of 113.96 and standard deviation of 46.46. The autocorrelation of 1 h interval gas price averages indicates daily seasonality. A pre-processing approach of down-sampling gas price data to 5 min resolution, deletion of outliers above 2 standard deviations, and Fourier transform based denoising is employed [15].

Pierro [31,32] looked at Gas Oracles' forecasts, finding they are less accurate than claimed and user-defined categories for these prices are incorrect. To evaluate the accuracy of current Gas Oracles, the authors propose a user-oriented model based on two gas price categories that correspond to user preferences and a new way to estimate the gas price. Their method used Poisson regression at more frequent intervals, forecasting the price of gas with a narrower margin of error than the real one, giving users a more useful gas price to set.

Turksonmez et al. [33] developed a new gas prediction accuracy metric to assess oracle performance. They showed that oracles overprice transactions, leading them to reach the delay target but at a larger cost than necessary, as well as underprice transactions, causing them to miss the delay target. The authors compared five gas price oracles with results demonstrating relative accuracy, transaction accept rates, price stability, and discussion of factors that affect oracle accuracy. They noted that the ETHGasStation oracle generated the most precise and consistent pricing forecasts.

In an attempt to improve on oracle performance, particularly during times when transaction volumes are increasing rapidly and gas price oracles can underperform, Chuang and Lee [34] showed that Gaussian process models can accurately forecast the distribution of the lowest price in an upcoming block in the face of such increasing transaction volumes. Using the GasStation-Express and Geth gas price oracles, a hybrid model combining the two was proposed, providing a superior estimate when transaction volume fluctuates significantly.

Several modeling approaches are compared by Mars et al. [14]. Sliding windows of 300 previous blocks are used as input to forecast the next block ahead. GRU and LSTM models are found to have similar performance. Geth recommendations and Facebook Prophet forecasts are found to have similar performance, and they are outperformed by the RNN models. Down-sampling and outlier deletion pre-processing steps as found in Werner et al. are also employed before RNN modelling [15]. It is on the latter forms that we will concentrate for our study.

Laurent et al. provided a system of equations for calculating the probability a transaction is mined in a given period, given a gas price and knowledge of all transactions. The system was extended to predicting the probability of transactions being mined by the inclusion of a model for arrival or future transactions. The optimal gas price for a transaction to yield a specified probability of the transaction being accepted, in a given time frame, was achieved using a binary search of the transaction position within the set of modeled transactions. The authors state that comparison was difficult with previous works, as the probability estimate is a fundamentally different output to existing oracles or machine learning forecasts [35].

*4.2. Time Series Signal Processing and Data Mining*

Dyllon et al. demonstrates wavelet transforms for denoising and signal frequency-time density visualization. Of particular relevance is that a wavelet decomposition-based denoising approach is able to reduce noise in the high granularity, high noise signal while preserving seasonal elements. Continuous Wavelet Transform (CWT) is also used to visualize changing the signal frequency content over time [21].

Barry and Crane [36] showed that motifs and matrix profiles can be effective in improving the performance of LSTMs and the prediction of Bitcoin through the use of an LSTM neural network, yielding an 8% decrease in RSME for one test case. Sun and Xu applied wavelet coherence for the analysis of co-movement and lead–lag effect in multiple stock markets. Wavelet coherence allows a three-dimensional analysis of two signals on the axes of time, frequency and strength of correlation. Phase difference analysis is used to provide information on co-movement sign and lead–lag relationships [17]. Wavelets have application to a wide variety of time-series data, as shown by application to wide-band power signals [16], widespread use in financial market studies [17] and Geodetic signals [18].

*4.3. Deep Learning Models*

An Encoder–Decoder LSTM with attention guided by a matrix profile, as seen in Liu et al., can outperform other RNN models on low granularity data [19]. Fajge et al. [37] used a number of machine learning methods to determine if a transaction with offered gas fees is likely to be added to the blockchain within the anticipated period or not. Their results (evaluated on almost one million actual transactions from the Ethereum MainNet) showed that the proposed model outperformed existing ones at the time with an achievement of 90.18% accuracy and 0.897 F1-score when the model is trained with Random Forest on the dataset balanced with SMOTETomek. Qiu et al. apply an Attention-LSTM, with the degree of matching of each input element used to generate the attention distribution, and wavelet denoising is also applied; both wavelets and the attention mechanism improve performance compared to a standard LSTM [22].

CNN-LSTM models, LSTM models with pre-LSTM convolution filtering and feature pooling layers have seen widespread use in time-series forecasting. An attractive feature of CNN-LSTM models is the ability to effectively handle multiple inputs. Livieris et al. apply a CNN-LSTM toward gold price forecasting [38]. Widiputra et al. apply a single-headed convolution layer, fed into a two layer LSTM network, toward multiple output predictions of stock indices of Shanghai, Japanese, Singaporean and Indonesian markets [39].

Ferenczi and Bădică [40] investigated the prediction of Ethereum gas price with Amazon SageMaker DeepAR [41] and found that the choice of covariates had a large effect on model performance. They found that gas prices were impacted by various factors viz including seasonality, volume of transactions, transaction values, number of token transactions and amount of gas used per block.

*4.4. Research Gaps and Innovations*

We feel that our paper contributes to the literature through the following:

1. While a number of authors have covered the time period following the Ethereum London Fork (e.g., Refs. [26–28]), cited above, we feel that the relationship between EthUSD/BitUSD and gas price posited in Research Question 3 of our study provides an update on Pierro and Rocha's work of 2019 [24] on the link. This, we think, is an important addition to the corpus of research given the wide fluctuations in the price of cryptocurrencies.
2. Specifically investigating the performance of forecasts over different horizons. These time horizons are useful, as a user must select between these and potentially be penalized in terms of cost or missed transactions for choosing one over the other. There is thus a real cost penalty for the user in not choosing correctly here.
3. In our study, we use multiple approaches: a direct-recursive hybrid LSTM forecasting approach, inclusion of an attention mechanism with the matrix profile, as seen applied to low-granularity daily COVID data and also Convolutional Neural Networks (CNNs) fed to LSTM architectures or CNN-LSTMs.

In the case of matrix profiles, as noted above, this is the first incidence that we could find of this method used in gas price prediction. With the developing work on this method, we feel there is considerable potential for the method to be used to characterize patterns in gas price time series.

**5. Materials and Methods**

*5.1. Research Framework and Methodology*

The essence of the problem at hand is optimizing costs for a transaction sender. Senders are required to submit the price they pay per unit of gas with their transaction; the risks associated with under/overpaying lie with the sender. Oracles exist to recommend a gas price to address this risk; however, these are limited to simple heuristics.

Previous studies have attempted to improve upon these oracles with time-series forecasting-based approaches; however, these are limited to a short lookahead window. To our knowledge, existing recommenders and studies are limited to short lookaheads on the order of 5 min [14], a single block [25,29], or a handful of blocks [15].

This study seeks to provide insight into gas prices further into the future than existing oracles and studies. Knowledge of when gas prices will be low, or high, and the magnitude of these movements are proposed to provide value when planning transactions. For the purpose of generating this insight, the problem is framed as a time-series forecasting—supervised learning problem. Working within this framework is advantageous, as there as there are a wealth of available methods within this framework and a large body of existing work to draw from. LSTM models, attention models, and CNN-LSTM are all identified as powerful modeling approaches toward time-series forecasting. [20].

Time-series forecasting methods often make use of several data pre-processing methods before modeling. This study has identified wavelet transforms and the matrix profile as pre-processing and exploratory methods novel to gas price prediction and seeks to contribute understanding as to their applicability.

The presented methodology intends to investigate forecasting performance, across the identified modeling approaches, and pre-processing methods. Additionally, wavelet coherence is investigated as an exploratory tool.

*5.2. Description of Dataset*

Ethereum blockchain data were collected by query from the publicly available BigQuery database. Data spanning 26 November 2021 to 27 April 2022 were used in final modeling. Data were retrieved on a block-by-block basis with the final modeled dataset consisting of 953,336 blocks, averaging one block every 14 s, with an average of 203 transactions per block. Transactions were grouped by block to determine the block minimum, maximum and percentile gas price data, block transaction and contract counts. The gas used, base fee and size are provided on a block-by-block basis as is on the blockchain. The value of ETH cryptocurrency is known to affect gas prices [16]. Minute-wise tick opening prices of ETH, in US Dollar Tether, a stable coin tied to the price of the US dollar, were retrieved from Binance exchange historic records [42]. There were no missing data in the dataset.

*5.3. Wavelet Coherence*

Wavelet coherence is a bi-variate framework that probes the interaction of two time series on the basis of a wavelet function, over varying frequency scales, through time [17]. A wavelet $\psi$ is a time and frequency localized function with zero mean. The popular morlet wavelet can be defined as in Equation (1), with $\omega_0$ denoting the dimensionless central frequency.

$$\psi(t) = \pi^{-1/4} e^{i\omega_0 t} e^{-t^2/2} \qquad (1)$$

In order to compute the wavelet coherence spectrum, we first compute the Continuous Wavelet Transforms (CWTs) and cross-wavelet transform for the two time series. Equation (2) shows the CWT $W_x(\tau, s)$ of time series $x(t)$. The CWT is yielded by the inner product of $x(t)$ with a continuous family of "daughter wavelets" $\psi_{\tau,s}(t)$.

$$W_x(\tau, s) = \langle x(t), \psi_{\tau,s}(t) \rangle = \int_{-\infty}^{+\infty} x(t), \psi_{\tau,s}^*(t) dt \qquad (2)$$

Equation (3) shows the general form of a daughter wavelet. Daughter wavelets result from stretching the mother wavelet by varying $|s|$, and translating through time by varying $\tau$ with $s, \tau \in \mathbb{R}$, $s \neq 0$. Complex conjugation of the daughter functions is denoted by $\psi_{\tau,s}^*$. Varying $s, \tau$ in a continuous manner yields the set of daughter wavelets used in the CWT.

$$\psi_{\tau,s}(t) = |s|^{-1/2} \ \psi \frac{t-\tau}{s} \tag{3}$$

Equation (4) shows the cross-wavelet transform $W_{xy}(\tau, s)$, which can be defined in terms of the CWT of the investigated time series $W_x(\tau, s)$ and $W_y^*(\tau, s)$. These wavelet transforms can be interpreted as $\tau \times s$ matrices, indicating amplitude at scale s and time τ. $|W_{xy}(\tau, s)|$, the cross-wavelet power, indicates local covariance.

$$W_{xy}(\tau, s) = W_x(\tau, s) \ W_y^*(\tau, s) \tag{4}$$

Equation (5) shows wavelet coherence $R_{xy}^2(\tau, s)$, which can be estimated by using the cross-wavelet and auto-wavelet power spectrum, as laid out in Torrence et al. [43]. The $\tau \times s$ wavelet coherence matrix can be viewed as the time and frequency localized correlation of two time series on the basis of a wavelet convolution. The bi-wavelets package used in this project uses a modified version of this equation, as found in Liu et al. [44]. $S$ is a smoothing operator, which is achieved by convolution in time and scale.

$$R_{xy}^2(\tau, s) = \frac{\left|S\left(s^{-1}W_{xy}(\tau,s)\right)\right|^2}{S(s^{-1} \ |W_x \ (\tau,s)|^2) \ S(s^{-1} \ |W_y \ (\tau,s)|^2)} \tag{5}$$

Coherency does not distinguish between positive and negative correlation due to the squaring of terms. Equation (6) shows phase difference, incorporating the imaginary $\Im$ and real $\Re$ parts of the of the power spectrum, which can be used to differentiate between these movements and give information as to the leading/lagging nature of the correlation.

$$\phi_{xy} = tan^{-1}\left(\frac{\Re\{S(s^{-1}W_{xy}(\tau,s))\}}{\Im\{S(s^{-1}W_{xy}(\tau,s))\}}\right) \tag{6}$$

A phase is visualized as arrows in high-correlation regions. Right-pointing arrows indicate signals in phase, while left-pointing arrows indicate signals in anti-phase. The lead–lag relationship is indicated by arrows pointing right–up for first variable leading and left–down for second variable leading.

*5.4. Wavelet Denoising*

The Discrete Wavelet Transform (DWT) is a variation on the wavelet transform that uses a discrete set of mutually orthogonal wavelet scales as opposed to the continuous set found in the CWT. DWT decomposition is typically achieved through use of high/low-pass convolution filter banks. These convolution filters are designed using wavelet basis functions with perfect reconstruction and no-aliasing constraints. High- and low-pass filters are applied to the input to generate Detail $D_j$ and Approximation coefficients $A_j$. The same filters can be recursively applied to these coefficients to yield additional decomposition levels, as shown in Figure 2 [45]. DWT can be used to denoise specific frequency bands of a signal by applying a threshold to particular decomposition levels. The signal can then be reconstructed from the thresholded decomposition coefficients, using an additional set of inverse filters, which are orthogonal to the decomposition filters [21]. A hard thresholding approach is used, where values in decomposition level $D_j$ that are below a threshold $u$ are set to zero. Equations (7)–(9) show calculation of the threshold $u$ for a given decomposition level $D_j$. The threshold $u$ is calculated based on the mean absolute deviation of the decomposition level $MAD(D_j)$, a user-defined denoising factor $\lambda$, and the number of time points in the decomposition level $\#D_j$.

$$MAD(D_j) = \frac{1}{n} \sum_{i=1}^{n} |D_{ji} - (\bar{D}_j)| \tag{7}$$

$$\sigma_{D_n} = \frac{1}{\lambda}(MAD(D_j)) \tag{8}$$

$$u_{D_j} = \sigma_{D_j}\left(\sqrt{3\ln(\#D_j)}\right) \qquad (9)$$

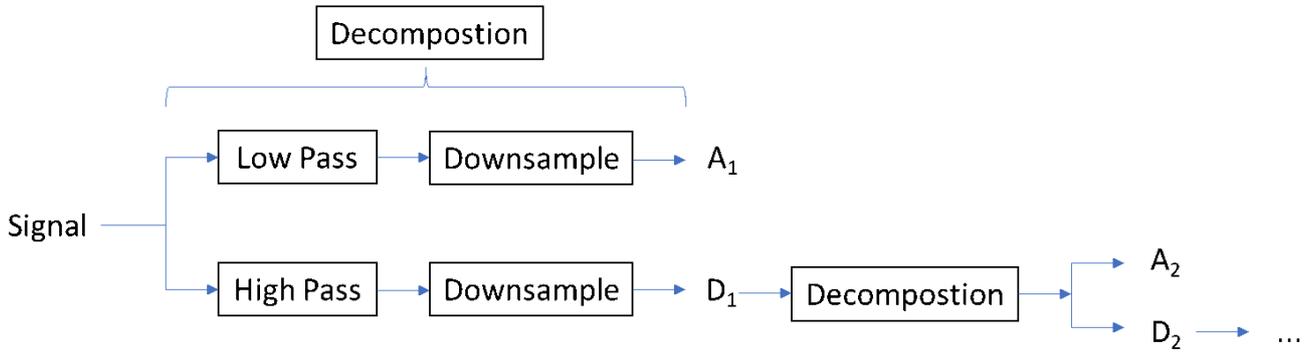

**Figure 2.** Wavelet Decomposition.

The effect of denoising can be evaluated by comparing the performance of models on raw vs. denoised data. However, this approach requires modeling for all denoising parameter sets to be tested, which is computationally expensive. The Signal-to-Noise Ratio (SNR) of the denoised signal and the RMSE of the raw vs. denoised signal can be used to indicate the effectiveness of the denoising approach and degradation of the signal, as seen in Qiu et al. [22]. Figure 3 shows how these RMSE and SNR measures are affected by altering the denoising factor $\lambda$.

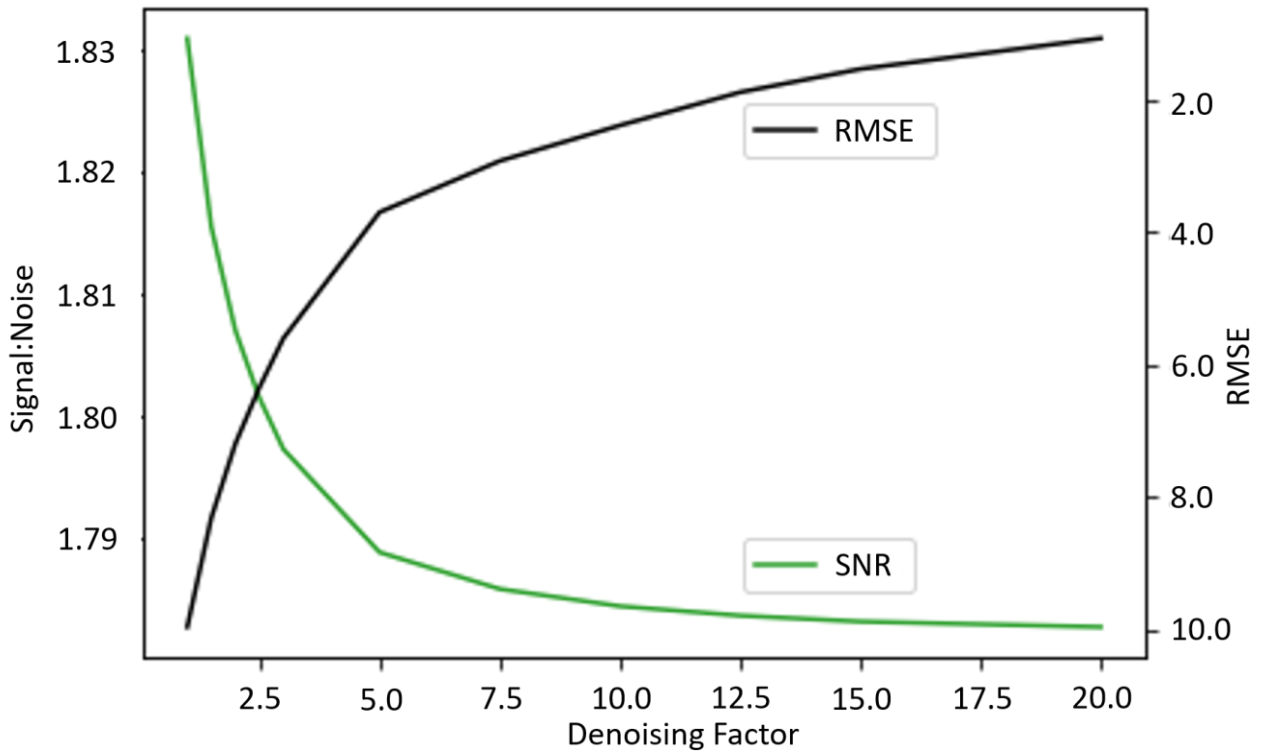

**Figure 3.** Evaluation of denoising of minimum block gas price using wavelet threshold denoising. Average RMSE and SNR of the top 5 wavelets by SNR shown across a range of denoising factor, $\lambda$ values. As $\lambda$ is increased, we see a decrease in RMSE and SNR.

*5.5. Down-Sampling and Normalization*

Data are down-sampled to the mean over a 5 min window before modeling. Initial modeling approaches truncated outliers in the block minimum gas price to a max of 2 standard deviations with min/max normalization. Z-score normalization, as shown in Equation (10), with no outlier truncation was used in later approaches. Z-score normalization of datapoint $x$ to $x'$ involves subtraction of the sample mean $\mu$ and division by standard deviation, $\sigma$.

$$x' = \frac{x - \mu}{\sigma} \tag{10}$$

*5.6. Matrix Profile*

The matrix profile is a companion time series [46] that indicates a similarity of subsequences in the parent time series. Given a subsequence size and input time series, the distance profile indicates the minimum Euclidian distance in terms of subsequence similarity to another subsequence of that size; points in the time series with a high matrix profile indicate the start of a discord, which is a subsequence with little to no repetition in the time series; and low values indicate a motif, which is a subsequence that repeats within the time series [46]. It has been shown [36] that motifs and matrix profiles can be effective at improving the performance of LSTMs.

The matrix profile is calculated for minimum gas price data, and used as an additional input in modeling, to indicate the proximity of the nearest discord. The matrix profile is calculated on a rolling basis as the forecasting windows move forward to prevent leakage into the training data. The matrix profile series is always one full window size shorter than the input data. One window size of data is removed from start of the gas price and other variables, after calculation of the matrix profile, to align the size of the inputs model. Figure 4 shows the minimum gas price for a training example with its companion matrix profile. The matrix profile foundation Python package is used for the computation of matrix profiles throughout with a window size of 1 day.

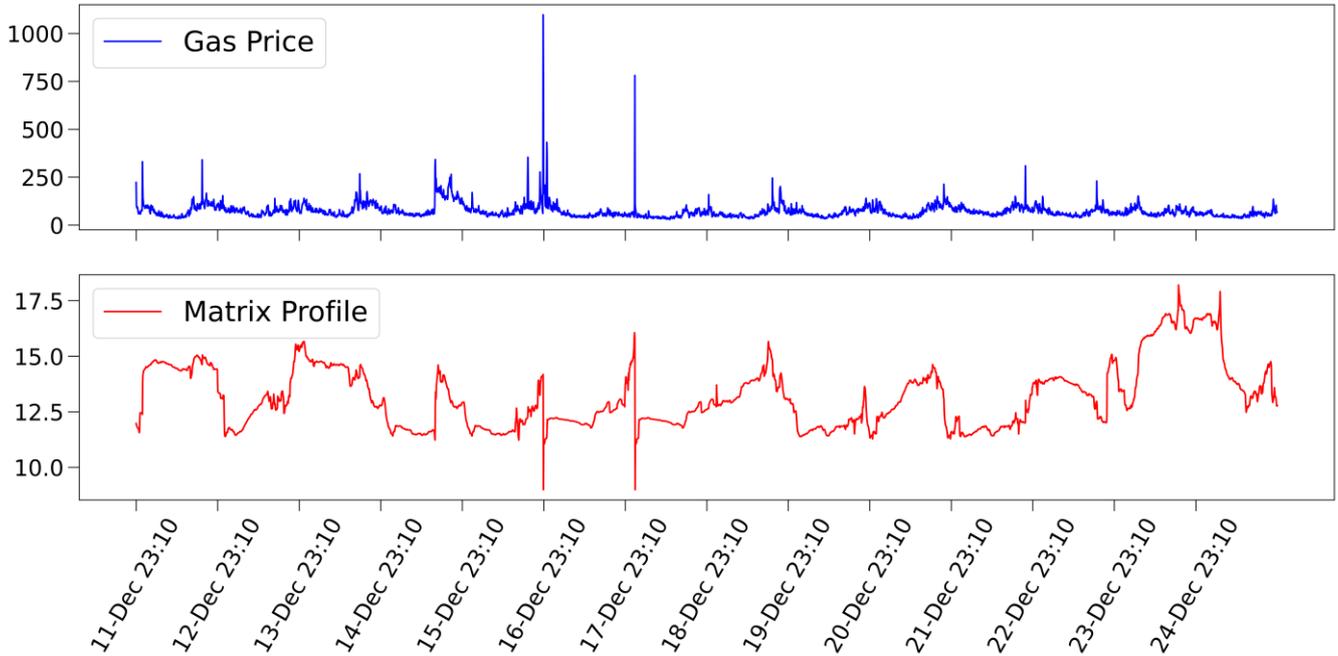

**Figure 4.** Minimum gas price and matrix profile.

## 6. Methods for Data Modeling

*6.1. Long Short-Term Memory (LSTM)*

Long Short-Term Memory (LSTM) features in a wide domain of time-series forecasting applications, including in stock market price prediction [22,38,39,47] and use in previous gas price forecasting studies [14]. LSTM networks were developed to address the problem of exploding and vanishing gradients in recursive neural networks (RNNs), particularly when information-carrying inputs are found several timesteps from the forecast window.

LSTM networks can be trained using a modified backpropagation algorithm, backpropagation through time, with gradient descent and its variations. The ADAM optimization algorithm is used throughout this project, and *tanh* activation is used in all cases to allow for GPU support with the cuDNN library.

*6.2. Recursive and Hybrid Strategies*

There are several forecasting strategies available for tackling the challenge of multi-step forecasting [48]. As shown in Equation (11), the recursive strategy first trains a model $f$ to predict one timestep ahead $y_{t+1}$ given an input series of $N$ observations; $t \in \{n, \ldots, N-1\}$. Extension of the forecast horizon past a one timestep lookahead is achieved by recursively appending the output to the input series and then feeding this new appended input back into the model.

$$y_{t+1} = f(y_t, \ldots, y_{t-n+1}) \qquad (11)$$

Equation (12) shows a direct strategy, which trains an independent model $f_h$ for each timestep in the lookahead horizon H; $h \in \{1, \ldots, H\}$

$$y_{t+h} = f_h(y_t, \ldots, y_{t-n+1}) \qquad (12)$$

Equation (13) shows a direct-recursive hybrid strategy, which combines the direct and recursive approaches. An initial model $f_0$ is trained as in the above models. A separate model $f_1$, as in the direct strategy, is then trained using the appended input of $f_h$, as in the recursive strategy. This process is recursively applied, learning H models $f_h$. This takes advantage of the stochastic dependency of the recursive approach while addressing its tendency for compounding errors with the direct multi-model approach. A hybrid LSTM model was trained to a lookahead of 10 timesteps, using hyperparameters from Mars et al. [13] for the base one-step lookahead model.

$$y_{t+h} = f_h(y_{t+h-1}, \ldots, y_{t-n+1}) \qquad (13)$$

Equation (14) shows a multiple output strategy. This strategy trains a single model $f$, which outputs $[y_{t+1}, \ldots, y_{t+H}]$ given $(y_t, \ldots, y_{t-n+1})$. This strategy approach allows for the modeling of dependency on future values, and it is a solution to impact stochastic dependency and compounding errors found in the single-output mapping models previously discussed. This is of particular concern when considering extended forecast horizons.

$$[y_{t+1}, \ldots, y_{t+H}] = f(y_t, \ldots, y_{t-n+1}) \qquad (14)$$

Multiple output models can be further applied in a direct-recursive manner. The $H$ step horizon can be segregated into several blocks. An initial model is trained to output the first block in the horizon; then, recursive training of new models with the inclusion of the previous models output as input is applied to generate the full horizon.

*6.3. Encoder–Decoder and Attention Mechanism*

Encoder–decoder networks function by first passing inputs into an encoder network. The encoder generates an intermediate representation of the inputs, which contains sufficient information for the decoder network to generate the target output. Encoder–decoder networks were originally developed to address sequence-to-sequence prediction problems in natural language processing, but they have since been widely adapted to time-series forecasting. The attention mechanism is a development that involves weighting outputs with an alignment of queries and keys [22]. A schematic of this is shown in Figure 5.

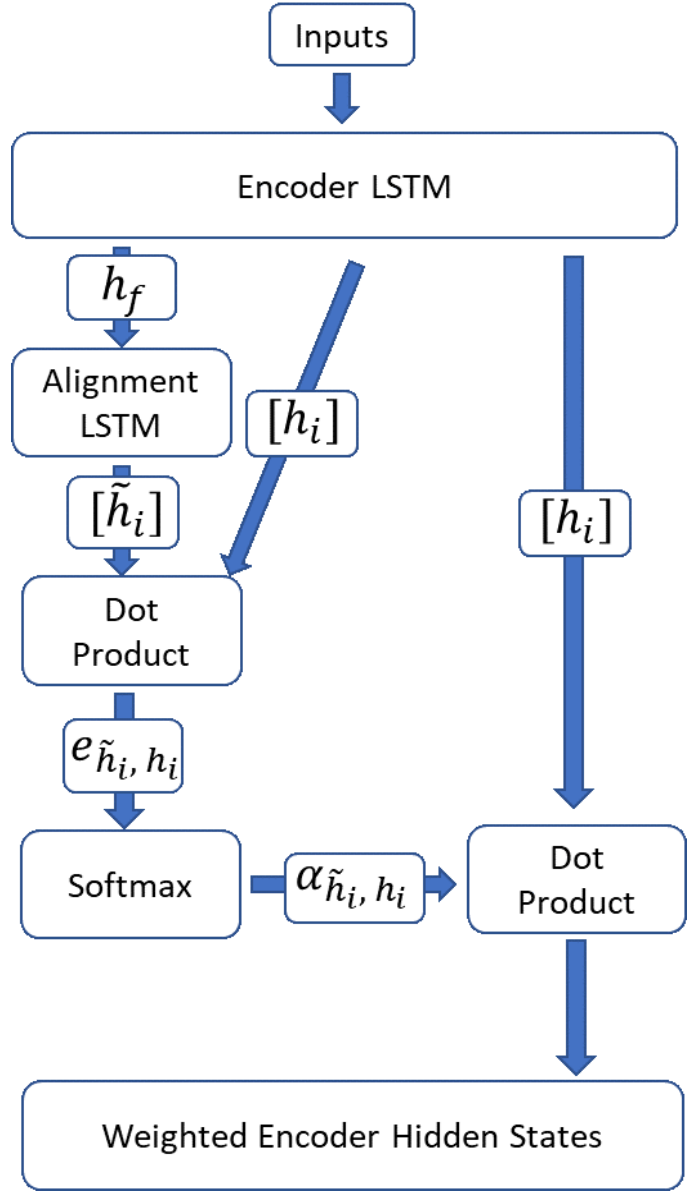

**Figure 5.** Schematic of Attention Head.

Equations (15)–(17) show the generalized attention mechanism, involving three primary components: queries $Q$, keys $K$, and values $V$. The dot product of query vectors $q$ and value vectors $v$, their alignment scores, is passed through a softmax activation to generate weights $\alpha_{q,k_i}$. The final attention score or Context vector $Context(q, K, V)$ is the sum of all weighted value vectors $\alpha_{q,\ k_i} v_{k_i}$.

$$e_{q,\ k_i} = q \cdot k_i \tag{15}$$

$$\alpha_{q,k_i} = softmax(e_{q,\ k_i}) \tag{16}$$

$$Context(q, K, V) = \sum_i \alpha_{q,\ k_i} v_{k_i} \tag{17}$$

Equations (18)–(20) show the attention mechanism used in this project. $k_i$ and $v_{k_i}$ are both set to the encoder hidden state at timestep $i$, $h_i$. $q$ is set to the hidden states of an alignment model $\tilde{h}$, which takes the encoder output $h_f$ as input. The mechanism essentially trains the alignment model to weight all hidden states $[h_i]$ of the encoder to generated the context vectors. These weighted hidden states are then passed to a feedforward layer to generate the forecast or to a second attention layer followed by a feedforward layer in the two-layer models.

$$e_{\tilde{h}_i,\ h_i} = \tilde{h}_i \cdot h_i \tag{18}$$

$$\alpha_{q,k_i} = softmax(e_{\tilde{h}_i,\ h_i}) \tag{19}$$

$$Context(q, K, V) = \alpha_{q,\ k_i} \cdot h_i \tag{20}$$

A multiheaded approach is applied, with one attention head, and one set of weighted hidden states being constructed for each input, with each head being fed all inputs. In the final model, these context vectors are concatenated and passed to a second layer of attention heads before passing this to a final linear layer. Both encoder and alignment LSTMs are set to 30 units each in the multiheaded models and 200 units each in the single-headed model. Training times were found to be similar for a multiheaded vs. single-head model with these hyperparameters of units.

### 6.4. CNN-LSTM

Convolutional Neural Networks (CNNs) consist of a bank of convolution filters and pooling layers. Convolutional layers function by scanning a convolution filter kernel across the data to generate new combined features. CNNs were originally developed for image processing, using two-dimensional filters. Multivariate time-series data can be treated in the same manner by representation in structured matrix form or by simply scanning a 1-D filter across each variable independently. Convolution layers are typically followed by a non-linear activation function, such as *tanh* or ReLU. Pooling layers then aggregate the output of convolution layer, typically taking the minimum, maximum, or average of a number of kernel outputs to generate a single datapoint [38].

A CNN-LSTM then passes the output of the CNN to an LSTM. This combined approach has seen much use in time series modeling, particularly financial data with complex multivariate dependencies. This project employs two layers of 1-D convolutional filters with a *tanh* activation function and no pooling layer. The final model uses a multi-headed architecture with independent convolution layers being fed all inputs. The number of heads is set to the number of inputs. After a grid search on the first month of data, a final filter size of 7 with nine filters per convolutional layer were used. These were fed to two LSTM layers of 100 units each.

### 6.5. Training Strategies

A sliding window of fixed input timesteps followed by a fixed number of forecast timesteps is used to generate training/validation examples. In all models, 70% of these examples are used for training, and 30% are used for validation. A walk-forward approach is desirable; however, the dataset contains over 40,000 timesteps even when down-sampled to a 5 min resolution. It is not feasible to walk-forward with every timestep. A daily walk is employed in the univariate, single-step lookahead analysis displayed below in Figure 6 In all other cases, a model, or set of models in the hybrid strategy, is trained and validated on one month of data with metrics averaged over 5 months. Models are trained for 15 epochs in all cases, with callbacks set to save weights for the lowest validation loss model during training.

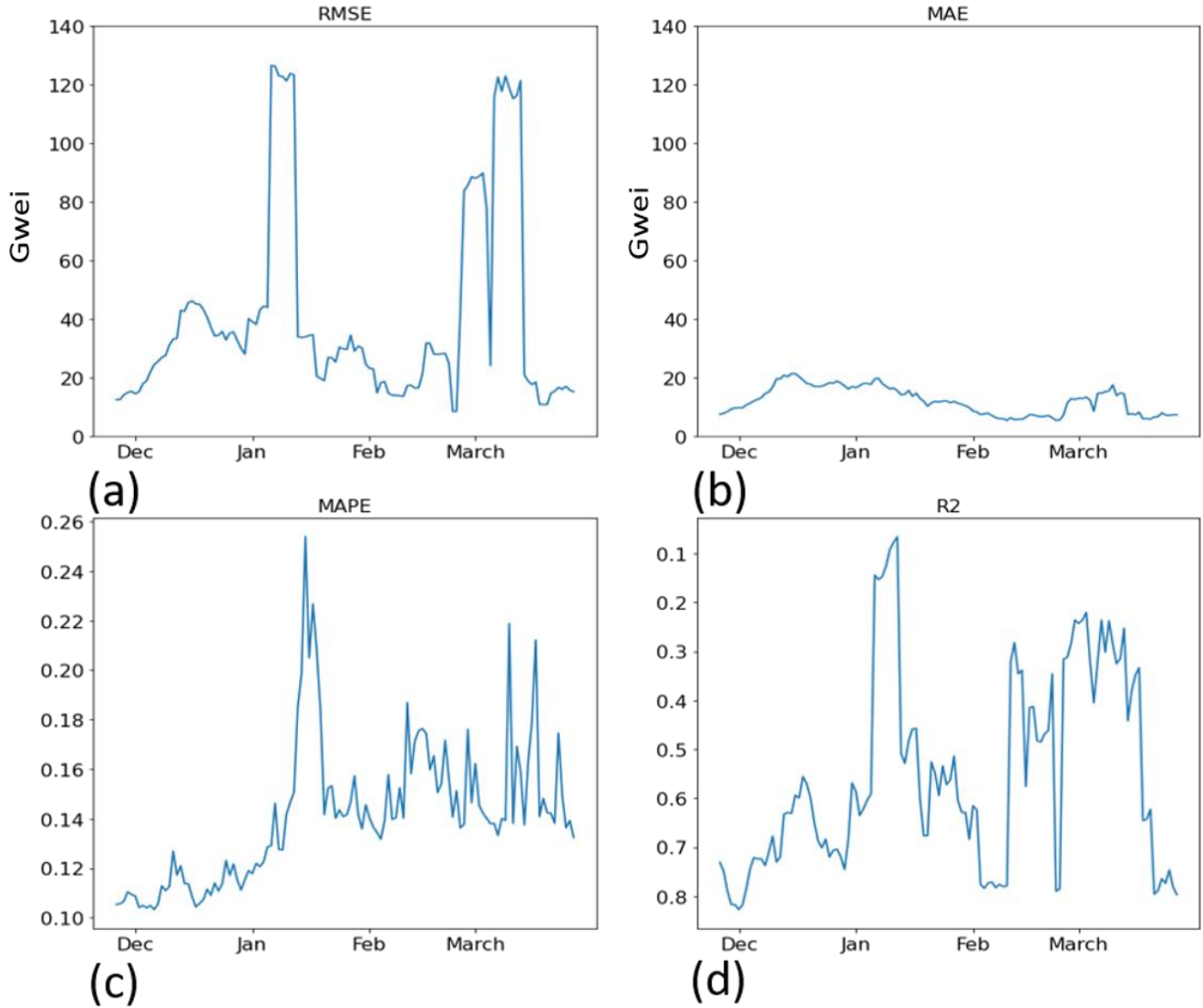

**Figure 6.** Univariate 1-step Walk-Forward Metrics. Blue lines refer to validation metrics for walk-forward, univariate, single-timestep lookahead model. Model is trained on one month of data validated with a 70:30 training: validation split; then, the data training/validation window was walked forward one day. *x*-axis represents start of the training period. (**a**) represents RMSE, (**b**) represents MAE, (**c**) represents MAPE, (**d**) represents $R^2$.

## 7. Results

### 7.1. Wavelet Coherence

Figure 7 shows coherence plots of block minimum gas price versus (a) Block Base Fee, (b) Gas Used, (c) Smart Contract-Type Transaction Counts, and (d) ETH/USDT ticker price. Base fee shows high correlation with signals in phase. Low-correlation areas can be seen in the time scale of 60–1000 min over narrow time periods. As can be seen from Figure 7a, volumes of high-priority, high-tipping transactions are sufficient to deviate the block minimum transaction gas fee selected by miners from the block base fee. Gas used, ETH/USDT and contract counts plots display noisy spectra at low time scales. Contract counts show strong anti-phase correlation at 1000 to 2000 min time scales across the majority of time periods. This is consistent with findings on a 1-day timescale in Donmez et al [24] regarding smart contract-type transactions being of lower urgency, and having lower gas price, than ETH transfers.

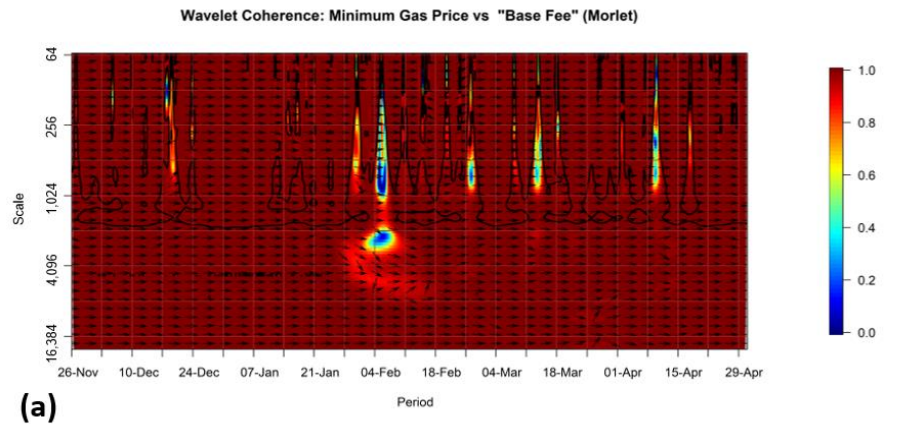

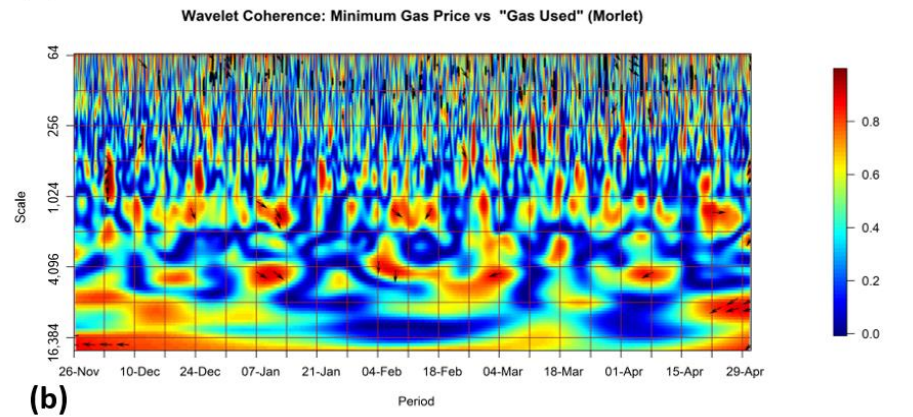

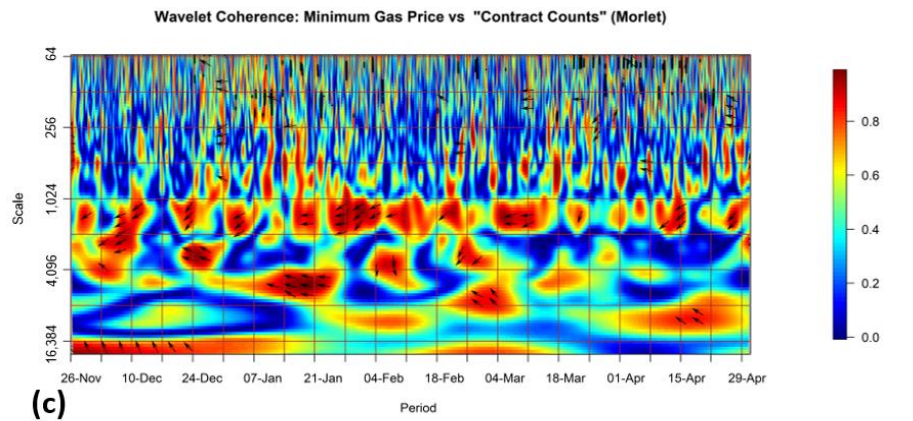

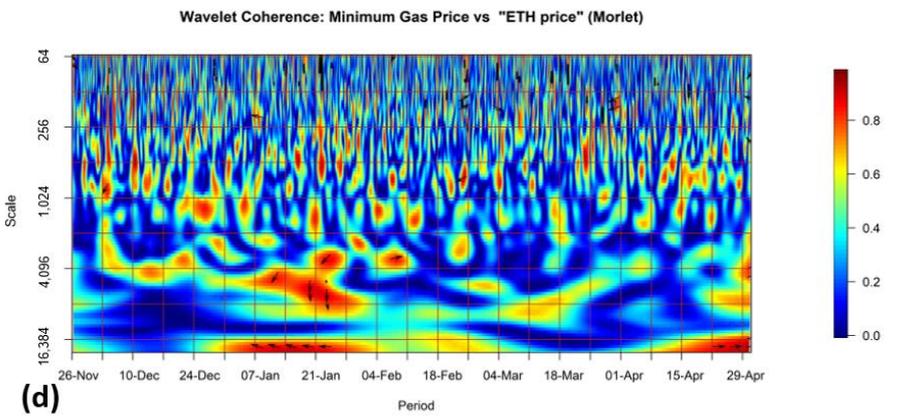

**Figure 7.** Wavelet Coherence Plots. Wavelet coherence plots of secondary variables against block minimum gas price. Time scale is in minutes. Period is in Month–Date. Data are down-sampled to a 5 min resolution before plotting. Heat indicates correlation and arrows indicate phase. Results show coherence plots of block minimum gas price versus (**a**) Block Base Fee, (**b**) Gas Used, (**c**) Smart Contract-Type Transaction Counts, and (**d**) ETH/USDT ticker price. Base fee shows high correlation with signals in phase. Low-correlation areas can be seen in the time scale of 60–1000 min over narrow time periods.

*7.2. Single Step Lookahead*

Figure 6 shows validation metrics for a univariate LSTM model, predicting one step ahead. The dramatic increase in RMSE and $R^2$ seen in Figure 6a,c, in windows starting in January and March is associated with extreme minimum gas price values in the validation data. This highlights the volatile nature of the data, sensitivity of metrics to changes in the data, sensitivity of metrics to changes in the data, and the need for a back-testing strategy to account for this behavior in the data.

Table 1 shows validation metrics for an LSTM model using minimum, 5th and 95th percentile gas prices, with additional variables. Hyperparameters as optimized by Mars et al. [14] were used for basic LSTM modeling. Increasing the depth or width of the network did not noticeably improve performance. Additionally, the 10th to 90th block gas price percentiles in increments of 10 were tested as inputs with marginal differences in metrics. This architecture is unable to model the complex dependencies between the variables or that the majority of variance in a one-step lookahead scenario is accounted for by minimum gas price variable.

**Table 1.** Multivariate Single-Lookahead LSTM Error Metrics. Validation metrics for multivariate, single-step lookahead LSTM models. Average of 5 models, each trained/validated on different month of data taken.

| Variable | RMSE | MAE | MAPE | $R^2$ |
|---|---|---|---|---|
| No Additional Variables | 20.28 | 10.50 | 0.142 | 0.680 |
| Block Size (Gas) | 19.18 | 9.55 | 0.125 | 0.715 |
| Base Fee | 19.89 | 10.28 | 0.132 | 0.693 |
| Transaction Count | 20.00 | 9.94 | 0.129 | 0.687 |
| Block Size (Bytes) | 19.96 | 10.16 | 0.133 | 0.687 |
| ETH/USDT | 20.14 | 10.42 | 0.135 | 0.685 |
| Average Gas Price | 20.11 | 10.46 | 0.142 | 0.683 |
| Maximum Gas Price | 20.42 | 10.75 | 0.140 | 0.674 |
| Smart Contract Count | 20.09 | 10.40 | 0.135 | 0.684 |
| All of Above | 19.35 | 9.74 | 0.126 | 0.711 |

*7.3. Hybrid Models*

Direct-recursive hybrid strategies were employed with univariate and multivariate models. The base one-step lookahead models in the multivariate test showed poor performance metrics on variables aside from the minimum gas price. The ability to accurately predict multiple outputs indicates potential avenues for the development of a multivariate approach.

Tables 2–5 show the performance metrics for hybrid and multiple output models. Modeling strategies are applied to five separate one-month blocks of data; then, monthly metrics are averaged to yield Tables 2–5. Figure 8 shows RMSE and $R^2$ degradation as the lookahead horizon is extended.

**Table 2.** Hybrid Model: Average of 5 Lookaheads *, All Months. Validation metrics for multivariate, single-step lookahead LSTM models. Average of 5 models, each trained/validated on different months of data taken.

| Variable | RMSE | MAE | MAPE | R² |
|---|---|---|---|---|
| Att 1 Head | 27.15 | 15.89 | 0.226 | 0.435 |
| Multi-Att 1 Layer | 28.46 | 15.86 | 0.245 | 0.389 |
| Multi-Att 2 Layer | 24.70 | 14.00 | 0.199 | 0.521 |
| Multi-Att 2 Layer MP | 25.63 | 14.33 | 0.206 | 0.486 |
| Multi-Att 2 Layer Uni | 25.74 | 14.47 | 0.190 | 0.484 |
| Multi-Att 2 Layer Uni MP | 27.38 | 15.76 | 0.220 | 0.421 |

* Model parameter shorthand: Att → Attention; Multi → Multiheaded; MP → Matrix Profile; Uni → Univariate; Rev → MP fed in reverse; DB4 → DB4 denoised gas price; Bior 3.3 → Bior 3.3 denoised gas price.

**Table 3.** Hybrid Model: Average of 10 Lookaheads *, All Months. Validation metrics for multivariate, single-step lookahead LSTM models. Average of 10 models, each trained/validated on different month of data taken.

| Variable | RMSE | MAE | MAPE | R² |
|---|---|---|---|---|
| CNN | 27.30 | 16.25 | 0.230 | 0.414 |
| CNN MP FWD | 27.68 | 16.42 | 0.238 | 0.414 |
| Multi-Att 2 Layer | 27.00 | 15.60 | 0.217 | 0.436 |
| Multi-Att 2 Layer MP | 28.27 | 17.30 | 0.237 | 0.402 |
| Multi-Att 2 Layer MP DB4 | 27.13 | 15.37 | 0.213 | 0.435 |
| Multi-Att 2 Layer Uni Bior 3.3 | 27.85 | 16.38 | 0.232 | 0.410 |
| Hybrid | 26.08 | 13.09 | 0.171 | 0.5421 |
| Hybrid MP | 27.02 | 14.29 | 0.195 | 0.5166 |
| Hybrid MP DB4 | 27.27 | 14.34 | 0.193 | 0.5082 |

* Model parameter shorthand: Att → Attention; Multi → Multiheaded; MP → Matrix Profile; Uni → Univariate; Rev → MP fed in reverse; DB4 → DB4 denoised gas price; Bior 3.3 → Bior 3.3 denoised gas price.

**Table 4.** Multiple Output Model: Average of 5 Lookaheads *, All Months. Validation metrics for multivariate, single-step lookahead LSTM models. Average of 5 models, each trained/validated on different month of data taken.

| Variable | RMSE | MAE | MAPE | R² |
|---|---|---|---|---|
| Multi-Att 2 Layer MP Rev | 25.07 | 14.02 | 0.193 | 0.509 |
| Multi-Att 2 Layer Uni MP Rev | 25.54 | 14.17 | 0.194 | 0.501 |

* Model parameter shorthand: Att → Attention; Multi → Multiheaded; MP → Matrix Profile; Uni → Univariate; Rev → MP fed in reverse; DB4 → DB4 denoised gas price; Bior 3.3 → Bior 3.3 denoised gas price.

**Table 5.** Multiple Output Model: Average of 10 Lookaheads *, All Months. Validation metrics for multivariate, single-step lookahead LSTM models. Average of 10 models, each trained/validated on different month of data taken.

| Variable | RMSE | MAE | MAPE | R² |
|---|---|---|---|---|
| Multi-Att 2 Layer MP Rev | 26.78 | 15.49 | 0.221 | 0.452 |
| Multi-Att 2 Layer MP Rev DB4 | 26.82 | 15.17 | 0.212 | 0.450 |
| Multi-Att 2 Layer MP Rev Bior 3.3 | 27.25 | 15.65 | 0.228 | 0.431 |
| Hybrid MP Rev | 27.33 | 13.92 | 0.184 | 0.509 |
| Hybrid MP Rev DB4 | 27.40 | 13.82 | 0.179 | 0.508 |

* Model parameter shorthand: Att → Attention; Multi → Multiheaded; MP → Matrix Profile; Uni → Univariate; Rev → MP fed in reverse; DB4 → DB4 denoised gas price; Bior 3.3 → Bior 3.3 denoised gas price.

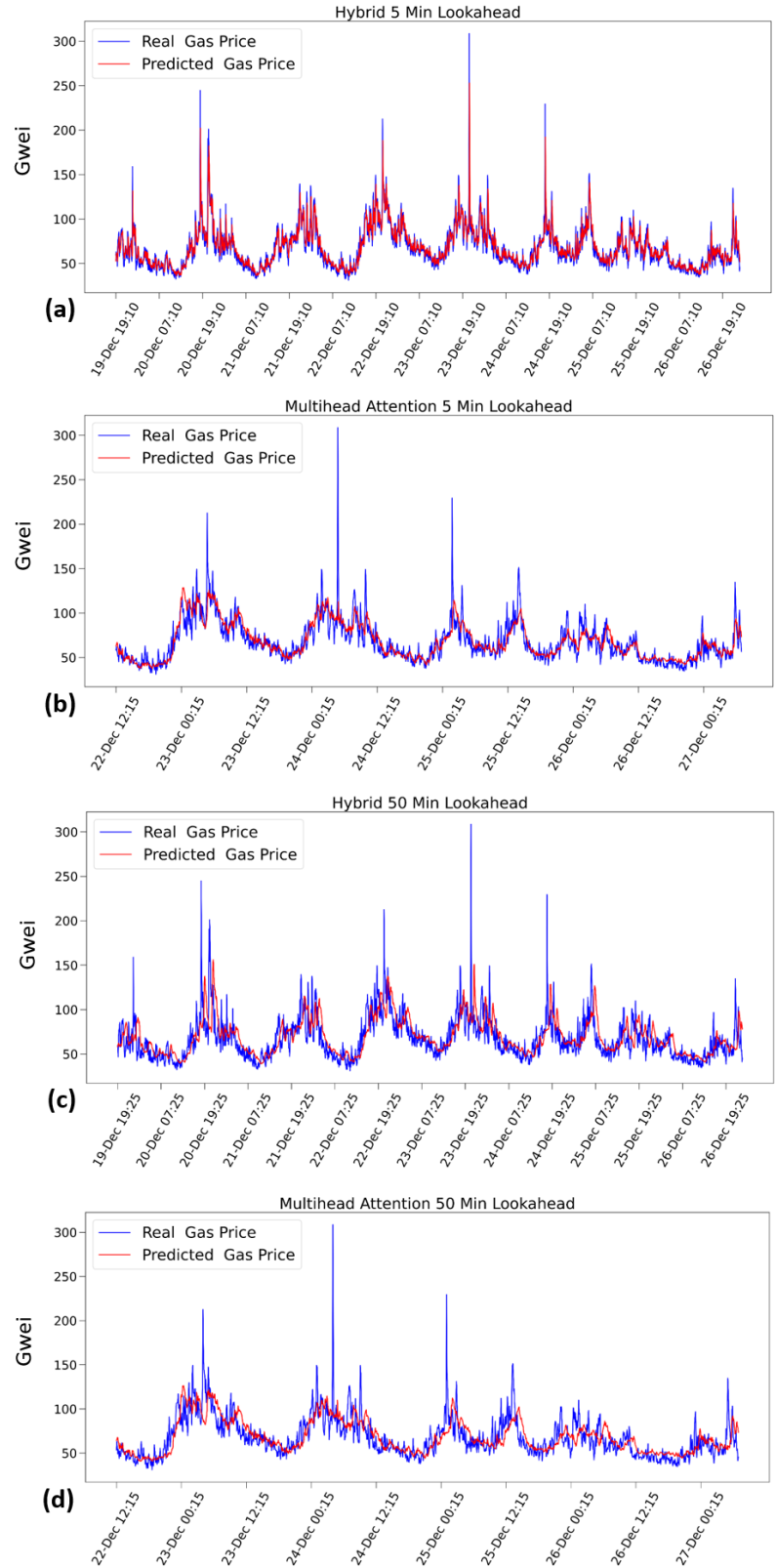

**Figure 8.** Validation Forecasts for Different Methods and Lookahead Window Lengths from 5 to 50 min (**a**) Hybrid 5 Min Lookahead; (**b**) Multihead Attention 5 Minute Lookahead; (**c**) Hybrid 50 Min Lookahead; (**d**) Multihead Attention 50 Minute Lookahead. Gas price values are quoted in gwei.

### 7.4. CNN-LSTM

Results of the grid search of multiheaded models found nine filters with a kernel size of seven to have the lowest achieved validation loss on the first month of data. The final model was then trained with the same inputs as the attention models. Metrics were comparable to the attention models and inferior to the hybrid model. The use of two-dimensional convolution filters has seen use in previous works involving multivariate time-series data, and we would recommend their investigation in future works [38].

*7.5. Attention*

The attention models were trained with a single-headed architecture, a multiheaded architecture, one and two attention layers, with a wavelet and matrix profile data preprocessing. The inclusion of multiple heads and multiple layers was found to improve validation metrics. Additionally, multivariate attention models showed better performance than univariate in contrast to the hybrid models. This may be because the more complex architecture is better suited to learning the complex dependencies between variables.

Figure 9 shows the performance of models at different lookaheads. Hybrid models were found to significantly outperform attention models at shorter lookaheads; however, only the univariate hybrid model has comparable metrics to the attention models at longer lookaheads. Averaging over all lookaheads, the best attention and hybrid models had similar RMSE; however, the hybrid model outperformed on other metrics. For reference, Figure 8 shows validation forecasts of hybrid and attention models for the 5 and 50 min lookahead. Comparing Figure 9a and 9b, respectively, it is evident that the hybrid model is much better able to track the stochastic movements of the data at the 5 min lookahead.

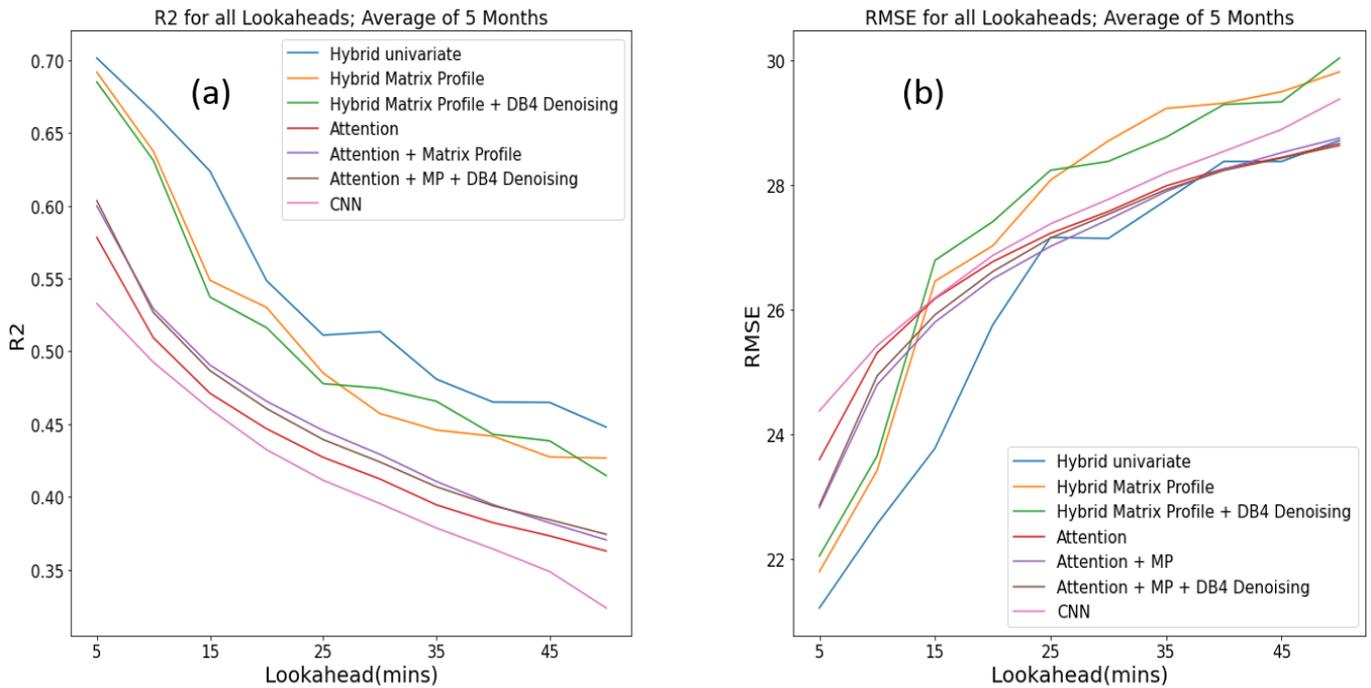

**Figure 9.** Performance Metrics ($R^2$ and RMSE) at different Lookaheads of the various models. (a) refers to $R^2$ metrics for different lookaheads and (b) refers to RMSE metrics for different lookaheads.

*7.6. Matrix Profile*

The matrix profile was fed as input to hybrid and attention models in both reverse and forward chronological order. In the case of hybrid models, addition of the matrix profile to the training examples had a negative effect on validation metrics. There is little difference in metrics between the reverse/forward matrix profile model in the hybrid case. This is likely to be due to the fact that hyperparameters had been tuned for a univariate model; the model may also not be sufficiently complex to extract the necessary features to make use of the matrix profile. We would suggest that optimizing the base one-step lookahead model with these inputs would be of interest in future works.

The addition of the matrix profile as an input to attention multiple output models showed inconsistent results. Inclusion of the forward matrix profile had a negative effect on the validation metrics in all attention models, and it had a marginal effect on the CNN model. Interestingly, the addition of the reversed matrix profile noticeably improved $R^2$ in the univariate five-step and all 10-step lookahead attention models as opposed to a decrease seen with the forward matrix profile. A reversed matrix profile did not improve metrics in the five-step lookahead multivariate model; however, the decline in metrics was less pronounced than with the addition of the forward matrix profile; the reversed matrix profile performed better than the forward in all attention models. This behavior could be explained by the introduction of some degree of bi-directionality into the models.

*7.7. Wavelet Denoising*

Wavelet denoising was applied with a Daubechies mother wavelet with scaling function 4, to the second decomposition level, $\lambda = 3$. Denoising with a biorthogonal wavelet with scaling functions (3,3) to the second decomposition level and $\lambda = 10$ was also tested, showing a noticeable decrease in $R^2$. The biorthogonal wavelet was selected as this wavelet provided the greatest signal to noise ratio gains at $\lambda = 10$, with RMSE of the denoised signal vs. the original of 2.37. The Daubechies wavelet was selected due to its popularity and use in wavelet coherence. In all cases, wavelet denoising was found to have marginal to negative effects on validation metrics. Between the selection of decomposition levels, mother wavelet, and thresholding parameters, the parameter space for wavelet denoising is considerable. A wider search of this parameter space would be of interest in future work.

Comparison with previous works is difficult; to our knowledge, no previous studies have attempted to forecast on a similar time scale. Additionally, the gas price optimization problem can be framed in a number of manners; as a forecasting problem, as a transaction selection probability estimate, or the various heuristic approaches found in existing recommenders/oracles. Future works could benefit from a reframing of the problem, such as applying machine learning toward a transaction inclusion probability estimate.

**8. Discussion**

*8.1. Research Questions*

In terms of Research Question 1, we found hybrid, multiheaded CNN-LSTM and attention approaches to be the best methods to forecast block minimum gas price. These were successfully applied to forecast multiple timesteps ahead, up to 50 min. The hybrid univariate model outperformed other models, particularly at earlier lookaheads. Attention models had comparable RMSE to the hybrid model at longer lookaheads but were outperformed on other metrics.

As regards Research Question 2, whether wavelet transforms and the matrix profile can improve forecasting metrics, or provide insight into gas price mechanics, wavelet denoised and matrix profile data were tested with a variety of modeling approaches, with mixed results. Wavelet denoising was not found to have any beneficial impacts on validation metrics; however, a narrow set of possible parameters was tested, so broad conclusions cannot be drawn as to the utility of the method in this domain. Matrix profile data fed in forward chronological order were not found to improve validation metrics in any models. However, interestingly, feeding matrix profile data in reverse was found to improve some attention models.

In order to answer Research Question 3, on the relationship between blockchain and ETH cryptocurrency exchange data on the one hand and gas price on the other, and whether these data can be used to improve forecasting metrics, we looked to wavelet coherence for insights. Wavelet coherence demonstrated a tendency for variables to correlate on a 1-day timescale. Smart contract counts were found to have strong anti-phase correlation on a 1-day timescale, which is in agreement with previous works [24]. Additionally, deviation of the base fee from the block minimum gas price can be seen at specific time periods, across a wide range of time scales, indicating periods of high numbers of high-priority transactions. Variability in univariate walk-forward metrics demonstrates the volatile and changing nature of the data, and it is an indication of the challenges modeling these data presents. The utility of additional variables beyond the gas price appears to be dependent on modeling architecture; additional variables had no effect on hybrid/one-step lookahead models but were beneficial in attention models.

*8.2. Comparison with Previous Works*

Comparison with previous works is difficult; to our knowledge, no previous studies have attempted to forecast on a similar time scale. Additionally, the gas price optimization problem can be framed in several manners; as a forecasting problem, as a transaction selection probability estimate, or the various heuristic approaches found in existing recommenders/oracles. Future works could benefit from a reframing of the problem, such as applying machine learning toward a transaction inclusion probability estimate.

Work by Mars et al. [14] is the most directly comparable. Mars and this work both operate with data down-sampled to a 5 min window, with Z-score normalization, within a supervised learning framework and with similar performance metrics. We can easily compare one timestep lookahead metrics; however, Mars do not provide forecasts past the first 5 min window. The authors in [14] provide MSE, MAE, RMSE and $R^2$ metrics, as found in this work. $R^2$ is most directly comparable as it dimensionless. Mars achieved an $R^2$ score of 0.896 on both GRU and LSTM-based forecasts, forecasting the minimum block gas price averaged for all blocks in the next 5 min. This study was able to achieve an $R^2$ of 0.715 within the same forecasting framework. The difference in performance can be attributed the more complete hyperparameter search performed by Mars et al. [14] and modeling on different time periods of data. MAE, MSE and RMSE metrics are quoted on a different scale to those found in this work, so a direct comparison is not possible.

Liu et al. [25] produce forecasts looking one block into the future. Forecasts presented by Liu achieve significantly better metrics than the models presented in this work that looked at 5 min windows; however, it is not clear how directly comparable these are given the timescale difference. This work was also completed before the London Fork. Liu et al. measured the proportion of their forecast values that falls into three categories: below lowest gas price and thus fail, above lowest gas price but below real gas price and thus succeed while saving costs, and higher than the real gas price, which succeeds but increase costs. This evaluation could prove useful in future works. Lan et al. [29] also produce similar one block ahead forecasts with improvements based on the addition of pending transactions in the Mempool as features. XG-boost based models outperform LSTM-based models in both cases and could be of interest in future forecasting studies on extended lookaheads.

Chuang and Lee [34] measure the performance of their model using two metrics: the first is success rate, or the proportion of their recommended transaction prices that are above the minimum gas price of the block, and the second is an Inverse Probability Weight measure (IPW), which increases with predicted gas price and decreases with success rate. IWP is used as the goal is to produce gas prices prediction that will result in successful transactions while keeping costs down. It is difficult to compare these metrics with those produced in this work. The success rate and IPW could be calculated for the forecasts generated in this work in future works.

## 9. Conclusions

### 9.1. Summary

In summary, this project has furthered forecasting attempts in an understudied area, with a novel combination of techniques, following a major update to the network in question. Gas price has been demonstrably forecasted at extended lookaheads; wavelet coherence has been shown to provide insight into the relation between gas price and blockchain variables, and the inclusion of matrix profile data showed a potential improvement of forecasting metrics. Direct Recursive Hybrid LSTM models were found to perform better than other modeling approaches given the limitations of the study. Further investigation is needed before drawing conclusions as to wavelet threshold denoising due to the limitations of the study.

### 9.2. Contributions

This study is the first that we have found to investigate gas price forecasting over different forecasting horizons. This study provides a methodology for forecasting gas prices up to 50 min ahead, in windows of 5 min. Forecasts over a range of lookaheads allow users to make an informed decision on gas price selection and the optimal window to submit their transaction in without fear of their transaction being rejected. This methodology provides more detailed and verbose information regarding gas price dynamics, in comparison to existing recommenders, oracles and forecasting approaches, that provide simple heuristics or limited lookahead horizons.

We have investigated multiple approaches toward generating the above-mentioned forecasts. Direct Recursive Hybrid LSTM models, attention models, CNN fed to LSTM architectures (CNN-LSTM), with matrix profile data and wavelet denoising were investigated. This is the first application of a matrix profile being applied to gas price data and forecasting that we are aware of. This study also demonstrated the applicability of wavelet coherence toward the analysis of movements in gas price data and related time-series data, insight regarding co-movements of gas price, block gas used, smart-contract transaction volume and ETH cryptocurrency price.

This study demonstrated that matrix profile data can enhance an attention-based model; however, given the hardware constraints, hybrid models outperformed attention and CNN-LSTM models. The potential for forecasting in extended and varying lookaheads was demonstrated with the utility of these time horizons being that a user must select between these and potentially be penalized in terms of cost or missed transactions for choosing one over the other.

The focus of this study was also to investigate data in the aftermath of the London Hard Fork, and it sheds insight into the transaction dynamics of the network after this major fork. We feel that this time period is of interest, as Research Question 3 of our study provides an update on Pierro and Rocha's work of 2019 [23] on the link between EthUSD/BitUSD and gas price.

### 9.3. Limitations of the Study

The limitations of this study are primarily related to available computing resources. Model training time was considerable on the available hardware. All data analysis, training and testing were performed on a desktop PC with a AMD Ryzen 5 1600 CPU and Nvidia 3060 GPU. The robustness of the training and testing strategy could be improved with a more thorough cross-testing method, such as a full implantation of walk-forward validation. The timespan of data considered is also limited due to the training time of models.

Optimizations of pre-processing methods and model hyperparameters were also restricted due to hardware limitations. A more thorough hyperparameter grid search or

Bayesian optimization would be of interest in future studies with more resources available. It is likely that the direct recursive hybrid model, an aggregate of many relatively simple models, outperformed the more complex CNN and attention models due to the above-mentioned limitations. Hybrid model hyperparameters are optimized for the single-timestep lookahead case; optimizing hybrid performance for longer lookaheads is also of interest.

The investigation of wavelet denoising and matrix-profile parameters were also limited by the model training time. The investigation of model performance when fed data using different wavelet-denoising approaches and different matrix profile window sizes and thresholds over different time periods with the varying model architectures previously mentioned would be of interest in future investigations.

*9.4. Future Work*

As mentioned in the limitations, future works would address limitations relating to the robustness of training/testing, time span of data investigated, and thoroughness of hyperparameter and pre-processing parameter search. Transformer models have shown promise in natural language and time-series forecasting problems; investigation would be of interest with sufficient resources for a through parameterization [49]. The XG-Boost-based model has also shown good performance in previous studies on this domain [25,29]. Several studies have investigated the use of Mempool data, investigation of these data toward improving forecasting performance and price dynamic understanding is also of interest [28,29,34,40].

Future works could also take advantage of domain-specific evaluation metrics such as those found in Chuang and Lee [34] and Liu et al. [25] to allow for better comparison of performance and more meaningful measures of performance.

To conclude, this project has furthered forecasting attempts in an understudied area, with a novel combination of techniques. Gas price has been demonstrably forecasted at extended lookaheads; wavelet coherence has been shown to provide insight into the relation between gas price and blockchain variables, and the inclusion of matrix profile data was demonstrated to show a potential improvement of forecasting metrics. Further investigation is needed before drawing conclusions as to wavelet threshold denoising.

**Author Contributions:** Conceptualization, C.B.; Data curation, C.B.; Formal analysis, C.B.; Funding acquisition, M.C.; Investigation, M.C. and C.B.; Methodology, M.C. and C.B.; Project administration, M.C. and C.B.; Resources, C.B.; Software, C.B.; Validation, M.C. and C.B.; Visualization, C.B.; Writing—original draft, C.B.; Writing—review and editing, M.C. and C.B. All authors have read and agreed to the published version of the manuscript.

**Funding:** For this research, the author M.C. wishes to acknowledge the support, in part, from the Science Foundation Ireland under Grant Agreement No. 13/RC/2106_P2 at the ADAPT SFI Research Centre at DCU. ADAPT, the SFI Research Centre for AI-Driven Digital Content Technology, is funded by the Science Foundation Ireland through the SFI Research Centres Programme. Both authors acknowledge the support of the Dublin City University Faculty of Engineering and Computing *Faculty Committee for Research* to meet publication charges.

**Data Availability Statement:** Data and implementation details are available at https://github.com/microlisk/Blockchain_Transaction_Fee_Forecasting (accessed on 01 May 2023).

**Acknowledgments:** The authors acknowledge the helpful suggestions from the reviewers and the assistance of the Editor. We also acknowledge permission granted by the authors in [14] to reproduce Figure 1 in their work.

**Conflicts of Interest:** The authors declare no conflict of interest.